\documentclass[a4paper,11pt]{article}
\usepackage{head,fullpage}     
\usepackage[pdftex]{graphicx}  
\usepackage{url}
\usepackage{datetime}

\newdateformat{monthyeardate}{%
  \monthname[\THEMONTH] \THEYEAR}

\title{\textbf{Using Deep Learning for Visual Decoding and Reconstruction from Brain Activity: A Review}}

\author{
        \textbf{Madison Van Horn} \\
        The University of Edinburgh \\
        Informatics Forum, Edinburgh, UK, EH8 9AB \\
        \texttt{M.A.Van-Horn@sms.ed.ac.uk} \\
}

\begin{document}
\date{}
\maketitle

\begin{abstract}
        This literature review will discuss the use of deep learning methods for image reconstruction using fMRI data. More specifically, the quality of image reconstruction will be determined by the choice in decoding and reconstruction architectures. I will show that these structures can struggle with adaptability to various input stimuli due to complicated objects in images. Also, the significance of feature representation will be evaluated. This paper will conclude the use of deep learning within visual decoding and reconstruction is highly optimal when using variations of deep neural networks and will provide details of potential future work. 
\end{abstract}

\section{Introduction}

The study of the human brain is a continuous growing interest especially for the fields of neuroscience and artificial intelligence. However, there are still many unknowns to the human brain. Recent advancements of the combination of these fields allow for better understanding between the brain and visual perception. This notion of visual perception uses neural data in order to make predictions of stimuli \cite{nestor}. The predictions of stimuli will allow researchers to not only see if we are capable of reconstructing visual thoughts but consider the accuracy with these predictions. This reconstruction is achievable due to the ability of using neural data from tools such as functional magnetic resonance imaging (fMRI) to assist in recording brain activity to later use for decoding and reconstructing a stimulus. These advancements enable the opportunity to further investigate how we can recreate perceived images and experiences by decoding human brain activity with the use of machine learning. 

As expected, issues arise with decoding visual stimuli. The type of input can have a great effect in the ability to accurately verify perceived images. For example, the use of natural images as input for neural decoding and visual reconstruction proves to remain a difficult task. The obvious solution is to choose wisely on the type of architecture to handle these complex, non-linear input data \cite{Livezey}. In order to solve this task of recreating images, deep learning has naturally transitioned into the field due to its success in other applications. This learning method is an increasingly popular choice in representing decoding and reconstruction models because of the great improvement and enhancement of image quality and reconstruction efficiency.

In this literature review, I will analyze recent developments of visual image decoding and reconstruction. These methods implement state of the art deep learning techniques such as deep neural networks, convolutional neural networks, variational autoencoders, and generative adversarial networks. In some cases, the same architecture is used for decoding and reconstructing which is known as an end-to-end model \cite{Shen_end, Han}. We are interested in seeing how these architectures can enhance image reconstruction since there are difficulties with achieving clear, quality images. I will also examine if there is a relationship between the architectures for the decoder and reconstructor as well as a relationship of homology and the process of visual image reconstruction. We want to learn about the impact of feature representations and homology. More specifically, the following research questions will be examined and investigated within this paper:

    \begin{itemize}
        \item What is the role of the architecture in the quality of the image?
        \item How well are these techniques adaptable to different domains?
        \item What is the significance of representation for features?

    \end{itemize}

The goal of this paper is to introduce and examine the current state of the art decoding and reconstruction techniques. Comparisons of the models will be made as well as present any discrepancies found during the analysis. This review will not cover the specifics of encoding models nor the fMRI analysis. As for fMRI, a discussion of this choice in recording brain activity will be presented with no further analysis of how the technique records data. Other brain activity recording techniques will not be examined including recording specific subregions of the brain due to being out of scope of the paper. The motivation behind this paper is to further understand how we can form a relationship between our brain and visual perception. This understanding of image reconstruction can lead to new societal impacts such as the ethics behind recreating seen images, dreams, and videos. 

This literature review will be structured as the following. Section 2 will provide background details to visual decoding and reconstruction with the use of fMRI data. Section 3 will contain the literature review with details of the role deep neural networks and generative networks play in the quality of the image, discuss the brittleness of the techniques applied to different domains and review the significance of feature representation. Section 4 will summarize and conclude the findings of the review as well as provide future work.

\section{Overview of Perceived Image Reconstruction}

The interest in expanding our knowledge with how the brain perceives information and images is an ongoing research problem within machine learning and neuroscience. The type and amount of data needed for the reconstruction is an issue within this area of research. Although there are successful models capable of identifying and reconstructing faces from fMRI data, different models are needed to address other images, such as natural images. Natural images remain a difficult problem within visual reconstruction due to the nonlinearity of the image. The amount of data is a constant obstacle within the field of machine learning, especially for accurately reconstructing images. Due to brain activity containing high dimensional data while having low number of samples creates an issue of properly predicting the reconstructed image during training \cite{Firat}. Usually, only one out of these two problems are addressed in reviews.

The ability to reconstruct images is feasible because of the use of fMRI. FMRI is a neuroimage analysis technique used to measure human brain activity by using fMRI BOLD signals to measure external stimuli \cite{chen}. It is a crucial imaging technique that has advanced the studies of recreating external stimuli seen from the brain. Although this paper will not provide further details on how the fMRI records or extracts the significant signals, it is important to mention fMRI data is the input data used for reconstructing images. This technique is advantageous by allowing non-invasive recording of the BOLD signals, recording the blood flow within the brain \cite{chen}. The noise produced remains an issue in this field due to the unknown of whether the noise is coming from fMRI data or from the model producing the noise itself. This unknown greatly affects the accuracy of all the phases including encoding, decoding, and reconstructing visually seen images.

\begin{figure}[h]
    \centering
    \includegraphics[width=10cm]{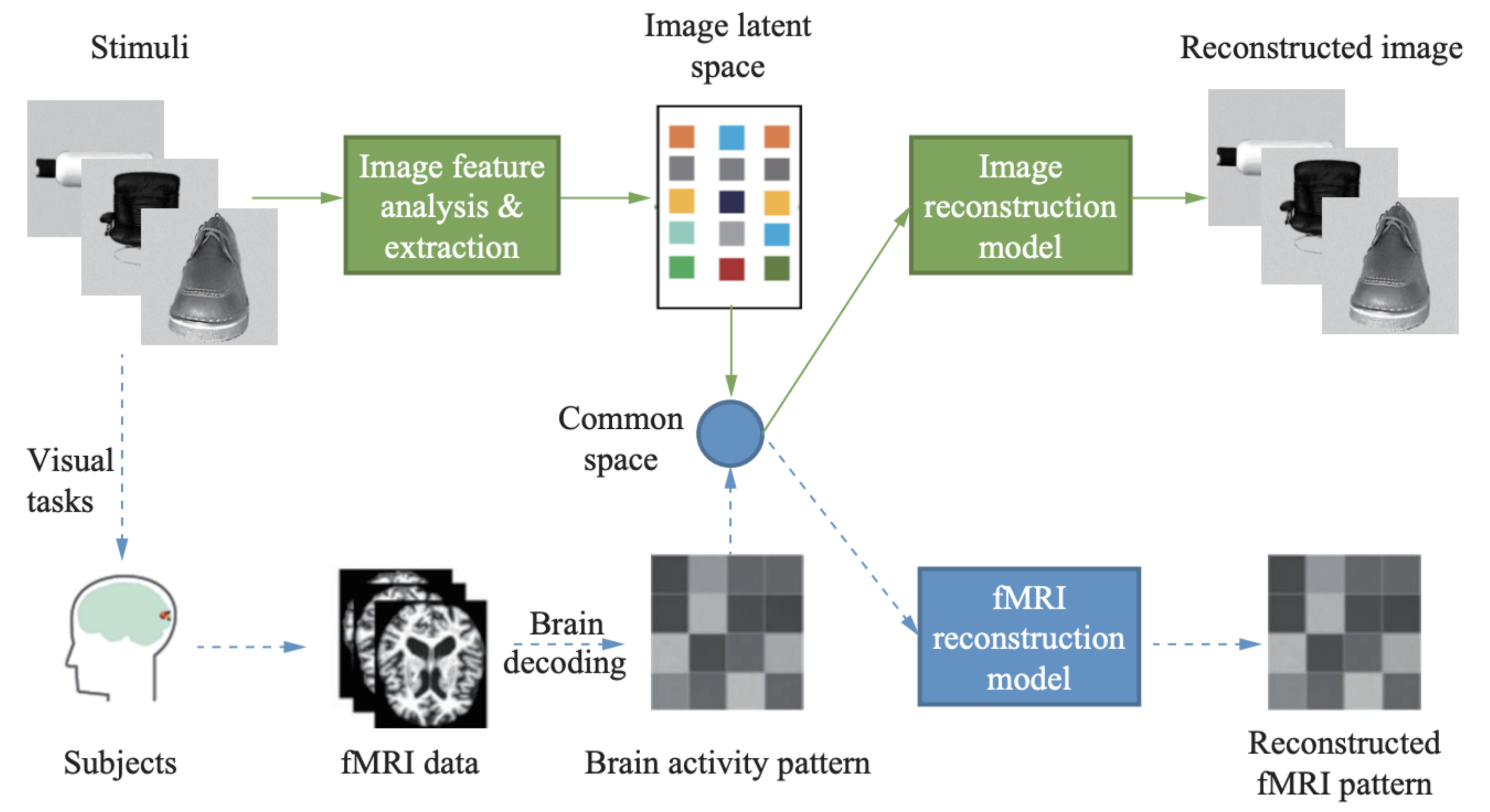}
    \caption{The above figure displays the process of brain decoding and reconstruction of images. \cite{shuang}}
    \label{fig:1}
\end{figure}

Visual reconstruction can be split into three phases: encoding, decoding, and reconstruction. As stated previously, this review will briefly give overviews of each phase but will only focus on decoding and reconstructing perceived images from the human brain. This work flow can be seen in Figure 1 where the process of brain decoding and reconstruction of images is shown. The encoding phase focuses on describing the individual voxels in a quantitative matter in order to keep important information about the stimuli. \cite{Naselaris}. This quantitative information provides descriptions of regions of interest in the human brain. The encoder must have the necessary visual features in order to provide decoders enough information to create quality images from stimuli. The decoding is the process of transitioning high dimensional brain activity into high dimensional stimuli \cite{vanGerven2019}. After retrieving the fMRI data, neural decoding occurs in order to understand the recorded brain activity data. This step is crucial in order to reconstruct the visual imagery process to achieve a quantitative expression from a phenomenological experience \cite{vanGerven}. Deep learning techniques can also be applied to neural decoders to achieve better decoding accuracy. The reconstruction phase uses this high dimensional stimuli in order to recreate perceived images from brain activity data.

The goal of visual reconstruction is to map brain activity to decoded features and to use these decoded features to reconstruct seen stimuli \cite{Wang_2020}. This is an important field of computational neuroscience to not only have a better understanding of how the brain processes information but how it perceives information or objects. The rest of the review will focus on the methods and domains applied to decoding and reconstructing images.

\section{Literature Review}
As this review will only focus on the decoder and reconstruction process of perceived images, we will elaborate on the role of the architectures allowing the ability to recreate and improve images. The model’s architecture plays a crucial role in the ability to show that newly created images are visually similar to the original image. It is evident deep learning has greatly expanded in the field of neuroscience and machine learning, allowing the ability to reconstruct perceived images from brain activity \cite{Firat}. Different deep architectures greatly contribute to the further development of visual reconstruction. We see the evolution of the various architectures and their roles in recreating the images have increased the quality of the reconstructed images to be similar in color, texture, and appearance to the original stimuli.

\subsection{Deep Neural Networks}

The emergence of deep learning greatly influences the model’s structure focusing on visualizing perceptual objects. Naturally, neural networks were first experimented to achieve this goal of image reconstruction, especially convolutional neural networks (CNNs). Shen experimented with implementing deep neural networks to perform deep image reconstruction from human brain activity \cite{Shen_deep}. The method pretrains 1,000 natural object categories with sixteen CNN layers and three fully-connected layers from ImageNet 2011 release. Effectiveness of the model was measured by pixel-wise spatial correlation and human judgment. The authors emphasize on correlating hierarchical visual features to the fMRI activity data due to similarity between hierarchical feature representation and the brain. The intentions were to optimize the image by making the deep neural network features similar to the decoded features. Although this network structure shows the capability of reconstructing images, it lacks quality and similarity to the original image. As the authors are advocates of generality, their model was only trained on natural objects which were centered. This is an issue when other natural images are used that can potentially have many objects in the image. Training objects centered in the middle of the image will not be optimal, leading to only specific types of images centered having better chances of being reconstructed. However, this initial implementation further increased interest in using deep learning for visual reconstruction. More recent publishings by Shen suggest further improvements of the model by creating an end-to-end structure to visualize perceptual content from brain activity \cite{Shen_end}. The goal of this model is focusing on improving the quality of the image by finding a direct mapping with a direct approach of fMRI activity to stimulus space. The belief of using this an end-to-end approach is believed to be due to potential information loss when performing the decoding of features. This method consists of three convolutional networks: one with three fully-connected layers with six upconvolutional layers, one with five convolutional layers with three fully-connected layers, and one with five convolutional layers as well as an average pooling layer and two fully-connected layers. The experiment consists of natural images of 150 object categories with 1,200 images for training and 50 object categories and 50 images for testing. Pair-wise spatial correlation and human judgment was used to evaluate the reconstructed images to the original. In comparison with the previous approach, Shen’s direct mapping of fMRI data increases accuracy and quality in the reconstruction. This shows that information loss occurring in an non-end-to-end model can lead to less quality in images. One advantage between the two methods is Shen’s 2019 research included an ablation study which is not used in any of the mentioned papers in this review. This ablation study goes through each layer of the model and removes a loss function to compare the reconstructions. This approach is useful in determining if loss functions are crucial for reconstructing an image and having it resemble the original.  The experiment showed removing the loss functions play significant roles in the quality representation of the image, indicating all the loss functions have a large effect in the reconstruction from direct brain activity to perception.

As mentioned previously, Shen was capable of creating an end-to-end network resulting in an increase in quality within the reconstructed images \cite{Shen_end}. Another researcher, Han, approaches a different version of an end-to-end model attempting to use variational autoencoders, VAEs, for decoding and reconstructing images and movies \cite{Han}. Although encoding is not a focus in this paper, it is important to note the team tried to expand the same implementation for encoding, decoding and reconstructing. The goal of the method is to use VAEs to generalize the decoding procedure. Their results show VAE is capable of regenerating natural images and movies. After training over 2 million natural images from ImageNet ILSVRC2012 dataset, their method indicates to be successful by reproducing images without losing significant color, structure, and other image qualities. This shows that a single structure is more than capable of reproducing competitive results compared to other models that use multiple structures to decode and reconstruct images. In comparison to Shen’s model, VAEs can be considered to outperform CNNs in the sense that VAEs capture color from fMRI data better. However, there is a tradeoff in the sense that VAE’s reconstructed images were far blurrier than CNN’s images. Also, it is interesting to note that Han only reported quantitative results for movies even though images and movies were used for training and testing of the model. Providing the statistics for the images would be useful when comparing against other end-to-end methods. Shen reports 62.9\% via structural similarity index measure whereas Han shows about 50\%. Although Shen’s model earns a higher value for accuracy, Han’s method is acceptable especially for videos.

\subsection{Generative Networks}

Recent developments within visual reconstruction have shown image quality improvements can be achieved with the use of generative networks. The presence of generative adversarial networks, GANs, have proved to be consistent in improving image quality especially with fMRI activity \cite{Yves}. This compatibility is successful due to GANs conditional distribution. This conditionality allows the model to focus on specific details within an image in hopes to focus on the same important details in the original image \cite{Yves}. More specifically, in order to achieve the highest quality and accuracy compared to the original image, techniques have been expected to incorporate 3 different deep learning methods for reconstruction. 

Addressing the quality of image can be achieved by conditional progressively growing generative adversarial networks. Huang creates this network with three deep learning techniques for reconstruction (CAE, LSTM, C-PGAN) \cite{Huang}. This architecture is trained and tested on a total of 2750 natural images from 10 object categories. Similarly, 50 images were used for the test set with the remaining 2700 images for training. An advantage to this approach shows it is not necessary to pretrain the network with images whereas DNN, GAN, and autoencoder networks pretrains data. The other models need the pretraining in order to produce high quality images. This methodology plays a significant role in image quality due to the progressive growth of the generator and discrimantor in the model which gradually adds new layers \cite{Huang}. This allows specific details to appear during training \cite{Huang}. Therefore, by using progressive growth and conditional distributions, this model achieves high accuracy as well as better image quality. 

Another use of generative networks to improve quality of reconstructed images is by using deep generative neural networks. This approach by VanRullen trains a VAE network using GAN procedure over a large dataset of 202,599 faces \cite{VanRullen}. This methodology produces quality results such as high accuracy and similarity between reconstructed and original image. The advantage of using this architecture allows the ability to easily decode unseen images while producing realistic, clear results. However, this approach uses a dataset of faces whereas previously mentioned models used natural images. This can potentially be an indication it will not generalize well to natural images. 

Seeliger presents a generative adversarial network for reconstructing images to achieve similarity with the original image \cite{seeliger}. This is achieved by using a deep convolutional generative adversarial model to learn the latent space of compressed data. The 1200 images consisting of 150 object categories were converted to gray scale. The argument is to test the capabilities of using this architecture to produce images. The goal of this experiment is to verify the capability of the model even if the quality of the image is compromised. Experiments consisted of using three different data sets of handwritten characters, masked natural images, and natural objects. However, an issue arose from the not generating images from certain samples, suffering from mode collapse. This causes the model to only learn kinds of images in a set. Mode collapsing can be considered to be stuck in local optimum and therefore not learning well from other kinds of images. Overall, the authors show the potential of creating images from GANs even with the disturbances encountered.

\subsection{Adaptability to Different Domains}

After examining the various deep learning techniques, it is evident there is a common theme of focusing on specific datasets. Limitations in the availability of fMRI data is one noticeable concern within this field. This is mainly due to the lack of availability of resources including limited stimuli to human participants in order to record data. Although there is a restriction on the amount of images to show as stimuli, it is important to recognize that a strong architecture should handle various forms of data in order to reconstruct images of faces, animals, landscapes, and various objects. We want to examine how or in some cases why these techniques are adaptable or brittle. These architectures tend to focus only on specific forms of reconstruction such as focusing on details or specific image types. In particular, most of the reviews focus on specific image types. Due to the difficulty specifically with reconstructing natural images, the reconstruction architecture’s adaptability appears to be brittle due to the lack of generalizing well \cite{Lin}.

The most common image types for visual reconstruction are facial and natural images. Although studies have shown facial reconstruction is capable of successfully reproducing quality images, more emphasis has been placed on expanding the quality \cite{Lin}. As can be seen in VanRullen’s work, a variational auto-encoder and generative adversarial network are used to reconstruct celebrity faces. This specific network successfully reconstructs facial images due to mapping the latent space and fMRI data instead of learning the pixels \cite{VanRullen}. Although facial reconstruction tends to be easier than reconstructing natural images, it is important to note this specific model provides adaptability for another domain such as gender recognition. This makes the model useful for different purposes other than just reconstruction. This method appears to be similar to Ren’s version of using dual variational auto-encoder and generative adversarial networks except for testing on natural images including handwritten numbers, characters, and colored photos \cite{Ren}. We will consider these architectures similar enough to compare with different objects due to using VAEs for decoding and GANs for reconstruction, as well as using the same metric for measuring similarity between original and newly reconstructed image, structural similarity index (SSIM). In this case, Ren’s reconstruction method is useful because not only does the researcher use various datasets but also makes comparisons to another VAE-GAN architecture, similar to VanRullen’s model. We can see that VanRullen produced 50.5\% accuracy between the original and reconstructed image on celebrity faces. However, Ren’s research shows the dual VAE-GAN model outperforms with 58.7\% accuracy on handwritten characters. It also shows the accuracy is 27.7\% for the VAE-GAN model on the same dataset mentioned in Ren’s experiments. This indicates that the same model, VAE-GAN, does not generalize well to various datasets. Therefore, this is an example of similar architectures that are brittle for different domains. 

This can also be seen between Güçlütürk’s and Huang’s models. Both architectures are very similar in structure yet use different input data as well as generate reconstructed images. Güçlütürk’s approach uses deep adversarial neural decoding (GANs) for faces \cite{gg}. Huang also uses a GAN model in order to reconstruct natural images \cite{Huang}. While there are slight differences in the models, such as Huang’s having a synchronized and progressively growing GAN and Güçlütürk’s iteratively minimizing the GAN’s lost function, we can still compare their results with different input images. Güçlütürk’s achieved 0.6512 with uncertainty of $\pm0.0493$ with Pearson Correlation Coefficient \cite{gg}. Huang earned slightly above 0.2 with uncertainty of about $\pm0.2$ \cite{Huang}. Although we notice Huang’s model received a smaller number for representing the correlation of the original and reconstructed number, it is important to consider a few things. One thing to notice is Huang used natural images which is a more difficult task. Another thing to consider is there is a larger number in uncertainty. If we take uncertainty into account, Huang’s model could achieve a higher number, one closer to Güçlütürk’s. Altogether, these models show similar architectures with the use of GANs to reconstruct images. However, there are significant differences in performance showing the models are brittle to certain input image types.

The lack of adaptability continues with natural movies. Natural movies are another difficult dataset to reproduce due to the difficulty of splitting and reconstructing the movie into natural scenes \cite{Wen}. The goal of presenting movies as stimuli is to recreate what a person is seeing since it will not always be in image form \cite{Wen}. Wen describes how convolutional networks are used to encode and decode fMRI responses as well as reconstruct the data. In this case, 972 natural movie clips were shown to participants in order for the fMRI to record the responses. A CNN model then uses the feature extracted data to reconstruct the natural movie frames. This paper is significant due to decoding and reconstructing natural vision within the same structure. This success in reconstructing natural movies is because of the model focusing on foreground and suppressed backgrounds \cite{Wen}. Although this research uses still images for training, testing of reconstruction of images was not prioritized. Also, it is mentioned the model needed a reduced number of categories in order to be applicable to movies \cite{Wen}. There has not been much research into producing quality natural movies from brain activity as well as images. Recent research does not include movie stimuli with the latest architectures.

\subsection{Feature Representation}

As the review will not cover the specific details of encoding, it is important to recognize the significance of feature representation. Feature representation is critical in reconstructing images due to containing the important information that represents the visual stimuli. The representation not only allows researchers to better understand the importance of feature vectors, but it also shows evidence of homology between the feature vector and the human brain architecture.

Hierarchical architecture is ideal especially for feature representation. The reasoning for this is due to neural networks and the brain structure having the similar hierarchical representation. Convolutional networks have this architecture and tend to be common representations for feature extractions. As can be seen in Wen’s research, the use of this type of network allows for decoding and categorical semantic representation \cite{Wen}. While the model is specifically intended for decoding, the categorical features can be embedded within the layers of the CNN. Thus, the embedding allows for generalizable semantic space. Another neural network architecture used for feature extraction can be seen within Lin’s research of using a CNN architecture \cite{Lin}. The output of the first fully connected layer proved to be the feature vector. This step is performed in hopes of capturing more identifiable features within the image since other representations focus more on capturing the structure of the image. Lin’s method of feature representation as well as Wen’s shows feature representation from layers is successful due to being similar to decoded fMRI. This similarity greatly improves the accuracy in the quality of the image as well as the decoding. 

Deep neural networks achieve this hierarchical representation because of the similarity in the architecture and brain representation as well. Since the goal is to have similarities between the network and brain, features are extracted into deep neural network feature values. Horikawa approaches this concept by using two different models, AlexNet and VGG19, to decode the fMRI data into deep neural network feature values \cite{horikawa}. This study shows using deep neural networks as feature representations shows evidence of homology between the brain and the decoding architecture. However, Horikawa also concludes this form of structure indicates unequal decoding representations. This is important to note because this is one of the main ideas behind using neural networks for feature representation. These explicit differences of the human brain and the network architecture goes against the idea of a successful reconstruction model having similar features of the brain and model. 

This form of neural networks is comparable to the previous research by Lin due to both using deep neural networks, VGG19, for feature representation. Lin notes that the use of neural networks to represent the features is wanted but need to avoid large feature size \cite{Lin}. The difference between the two feature extractor methods is that Lin uses the first layer as the feature vector whereas the output of each hidden layer in Horikawa’s research was used. This greatly reduces the size of features and dimensionality while still producing effective results. While both methods are similar in feature representation, Lin’s method proves to preserve the important visual information while reducing the number of features \cite{Lin}. Other research by Zhang suggests hierarchical feature representations are also effective in feature representation. Similarly to Lin’s, a convolutional neural network’s layers are used as feature representation from human brain activity \cite{CZhang}. This method uses eight layers where the output of each layer represents the image’s features. Interestingly, the lower layers output produced better accuracy amongst the features by Pearson Correlation Coefficient. This result appears to agree amongst the other variations in hierarchical feature representation indicating when using neural networks as feature extractors, all models use either the first few layer’s output as representation of the features. This indicates the more important features of the images are preserved in these lower layers. The previously mentioned methods of using neural networks for feature representation agree there is correlation between the brain and hierarchical features structure.

\section{Summary \& Conclusion}

We have discussed the advancement in techniques pertaining to visual reconstruction from brain activity data. It is evident that deep learning greatly enhances the ability to perform brain decoding and visual reconstruction because of architectural similarities between the brain and deep learning \cite{Firat}. The various architectures play a significant role in the quality of the reconstructed images. As it can be seen, deep learning greatly enhances this process. The use of deep learning, such as CNN, VAE, GAN, allows the ability to create models to extract the necessary information from fMRI data to use as features for the model. This process proves to be a difficult challenge especially for the use of natural images. We also examined the hardships of adjusting models to handle various types of input. This proves to be a common issue to reconstruct various types of data with the same model. Even if the data is in the same category, images and movies can play a significant difference. Lastly, we examine the significance homology plays within the human brain and the feature representation. 

Further research can be improved by testing models on other datasets along with the natural images. As the natural images are necessary to include due their difficulty in reconstructing, it is important to also include datasets such as facial images, to verify the model is generalizing well to various datasets. Many of the research papers focus only on one particular task usually referring to one specific type of dataset. Also, future research can be done to find metrics to measure the accuracy of a reconstructed image. Currently, there are no set metrics to use in evaluating a reconstructed image. Most papers will use Pearson’s Correlation Coefficient, SSIM, mean squared, and other metrics but there is no set guideline to measuring the accuracy. Most studies need human evaluation to determine the quality of the image. Further research into measuring the quality of a reconstructed image should be investigated. 

Another potential work can be pursued by making adjustments to the GANs procedure. As shown in this review, the use of GANs has greatly increased the quality in reconstructed images. Future work can consider improving this method by working on the precision and naturalness of the visual reconstructed product \cite{Qiao}. Research should be investigated to improve the decoding framework for better extraction of features. More emphasis can be made on the decoding architecture to include temporal and categorical information from fMRI activity. A recent paper investigated simultaneously extracting both types of information from fMRI activity and shows higher accuracy from the decoding framework \cite{Wang_2020}. Improving the decoding and reconstruction to accurately use vital features to reconstruct images will be the future for brain decoding. Many of the papers in this field focus on the overall accuracy of the reconstructed image versus the original. Improving the decoding accuracy could potentially improve the final accuracy of the reconstructed image. Altogether, the decoding step should be investigated to further see if it is in its highest optimal state to improve the reconstruction process.

\newpage
\bibliographystyle{unsrt}   

\small
\bibliography{main}       

\end{document}